\pdfoutput=1
\RequirePackage{fix-cm}
\documentclass[smallextended]{svjour3}       
\usepackage[numbers,sort&compress]{natbib}
\smartqed  
\usepackage{hyperref,url}
\usepackage{times}
\usepackage{xcolor}
\usepackage{soul}
\usepackage{epsfig}
\usepackage{graphicx}
\usepackage{subcaption}
\usepackage{amsfonts}
\usepackage{amsmath}
\usepackage{algorithm, algorithmic}

\usepackage[utf8]{inputenc}
\begin{document}

\title{State Distribution-aware Sampling for Deep Q-learning 
\thanks{The first two authors contribute equally.
}}


\author{Weichao Li         \and
        Fuxian Huang       \and
        Xi Li              \and
        Gang Pan           \and
        Fei Wu
}


\institute{W. Li \at
              College of Computer Science, Zhejiang University, Hangzhou, China \\
              \email{weichaoli@zju.edu.cn}           
           \and
           F. Huang \at
              \email{hfuxian@zju.edu.cn}
           \and
           X. Li \at
              \email{xilizju@zju.edu.cn}
           \and
           G. Pan \at
              \email{gpan@zju.edu.cn}
           \and
           F. Wu \at
              \email{wufei@cs.zju.edu.cn}
}


\maketitle

\begin{abstract}
A critical and challenging problem in reinforcement learning is how to learn the state-action value function from the experience replay buffer and simultaneously keep sample efficiency and faster convergence to a high quality solution. In prior works, transitions are uniformly sampled at random from the replay buffer or sampled based on their priority measured by temporal-difference (TD) error. However, these approaches do not fully take into consideration the intrinsic characteristics of transition distribution in the state space and could result in redundant and unnecessary TD updates, slowing down the convergence of the learning procedure. To overcome this problem, we propose a novel state distribution-aware sampling method to balance the replay times for transitions with skew distribution, which takes into account both the occurrence frequencies of transitions and the uncertainty of state-action values. Consequently, our approach could reduce the unnecessary TD updates and increase the TD updates for state-action value with more uncertainty, making the experience replay more effective and efficient. Extensive experiments are conducted on both classic control tasks and Atari 2600 games based on OpenAI gym platform and the experimental results demonstrate the effectiveness of our approach in comparison with the standard DQN approach.
\end{abstract}

\section{Introduction}
\label{intro}
Reinforcement learning (RL) is an important area of machine learning, concerned with how an agent learns through trial and error so as to maximize long-term rewards by interacting with an initially unknown environment. It has been successfully applied in various domains, such as robotic control\cite{kohl2004policy, ng2006autonomous, gu2017deep, antonelo2015learning, ma2013state}, game play\cite{silver2016mastering, foerster2016learning, silver2017mastering, lample2017playing, tuyls2006evolutionary,granmo2013accelerated}, health and medicine\cite{asoh2013application, soliman2014personalized, kusy2014probabilistic, nemati2016optimal, chitsaz2011software} and so on. One of the critical and challenging problems in RL, especially for the tasks with huge state space, is how to approximate the state-action values from the limited experiences and simultaneously keep sample efficiency and less computation. Here, each experience or sample is a state transition represented by a tuple of (state, action, reward, new state).

\begin{figure}[t]
  \centering 
    \begin{subfigure}[b]{0.49\textwidth} 
        \includegraphics[width=\textwidth]{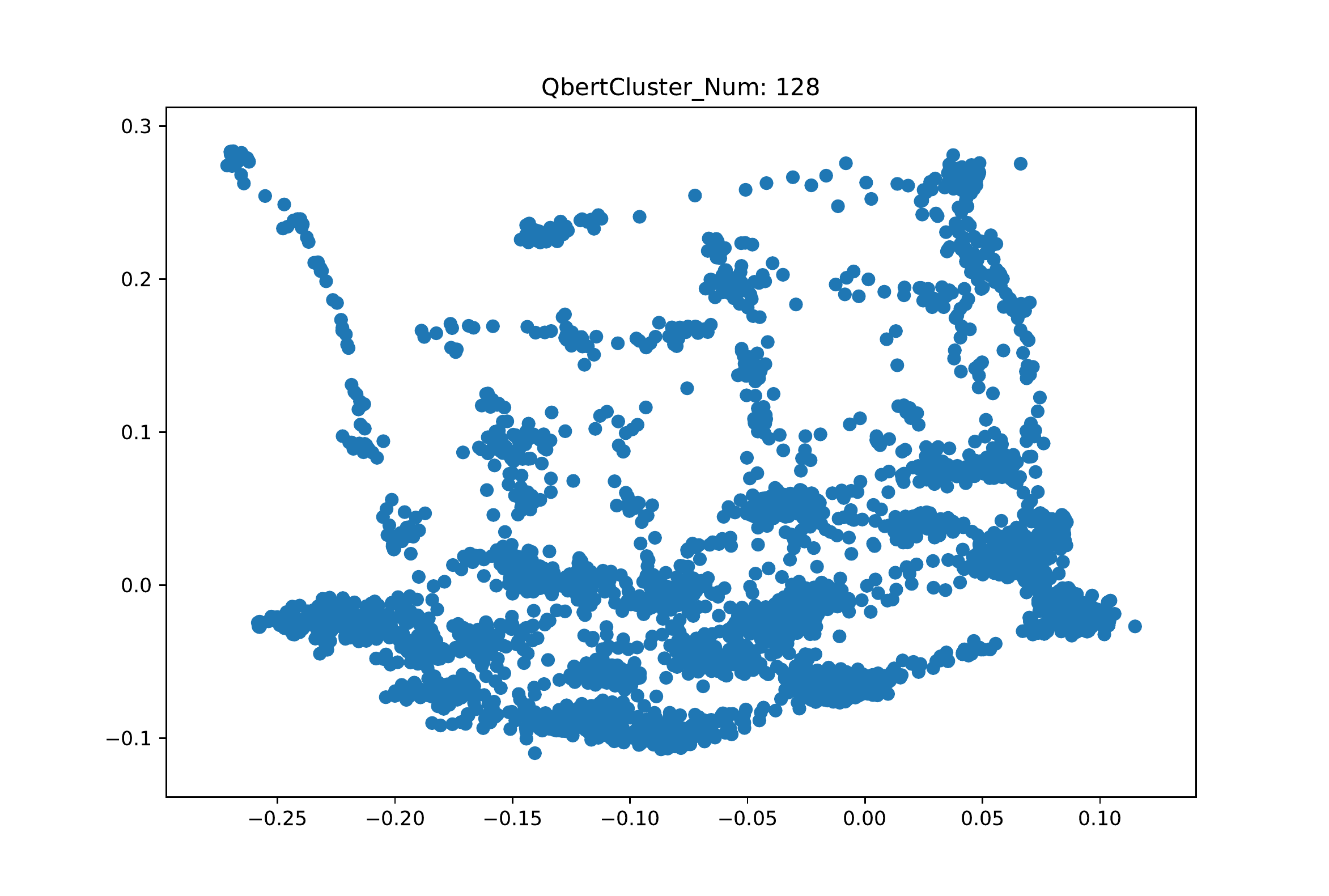}
        \caption{Distribution of transitions in state space}
    \end{subfigure}  
    \begin{subfigure}[b]{0.49\textwidth}
        \includegraphics[width=\textwidth]{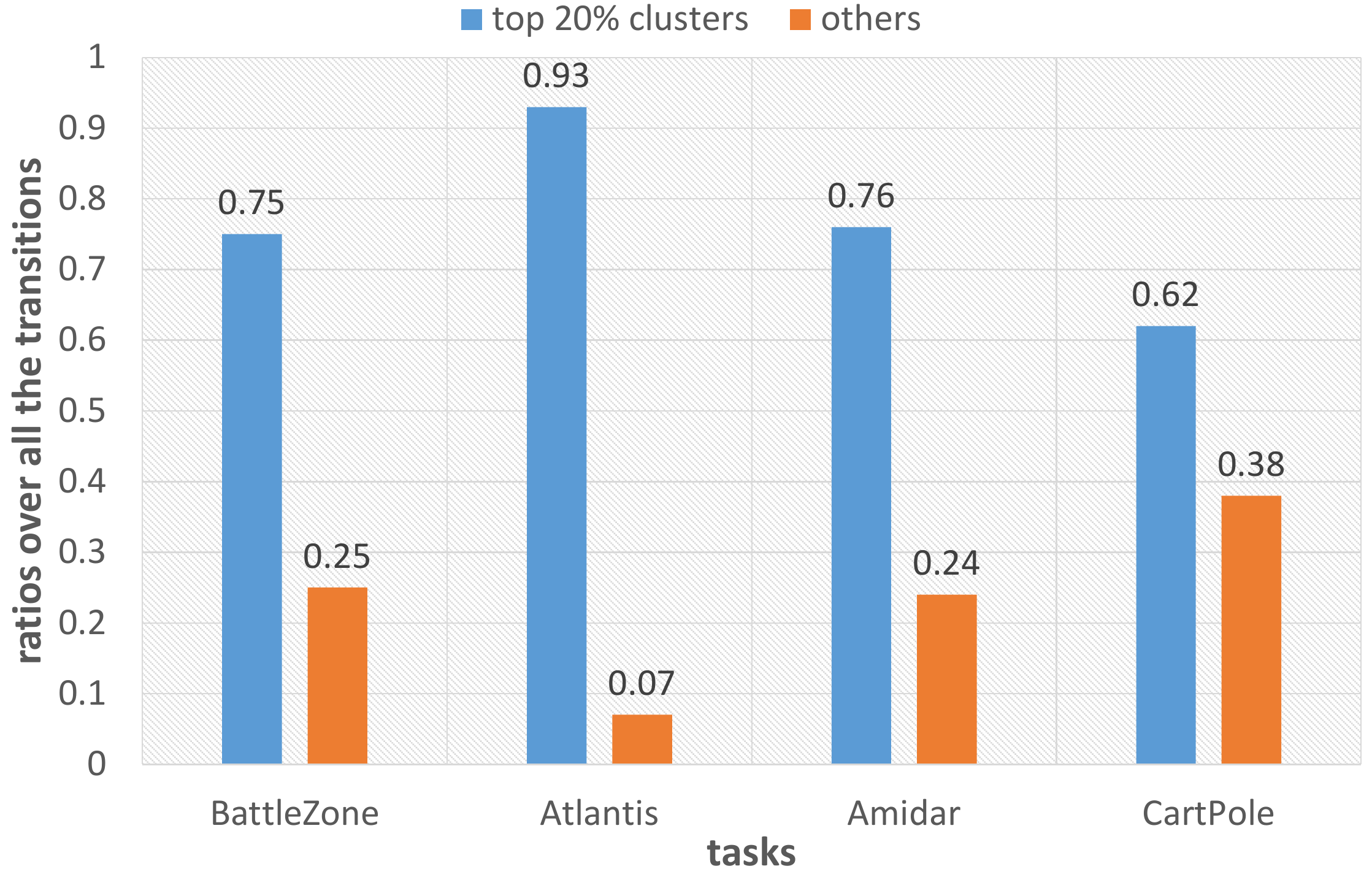}
        \caption{transitions ratio of top $20\%$ clusters}
    \end{subfigure}
  \caption{
    The transition distribution in replay buffer, which obtained by dimension reduction via PCA and t-SNE~\cite{maaten2008visualizing} in (a) and the percentage that the top $20\%$ clusters take up in the experience replay of four different tasks such as BattleZone, Atlantis, Amidar, CartPole in (b). It can see that more than $60\%$ of the transitions just belong to $20\%$ clusters. Therefore, the distribution of transitions is intrinsically skewed.
  }
  \label{fig:unbalance}
  \vspace{-1.0\baselineskip}
\end{figure}

Recently, it has become a trend to adopt deep neural networks combined with experience replay (ER)~\cite{lin1992self, mnih2013playing, mnih2015human, schaul2015prioritized, wang2016sample, silver2016mastering} to approximate the state-action values because of their good representation and generality. Generally, given the transitions $\{(s_{t},a_{t},r_{t},s_{t}^{'})\}_{t=1}^{n}$ ($s_{t}$, $a_{t}$, $r_{t}$ and $s_{t}^{'}$ represent state, action, reward and new state respectively), the state-action values are usually learned by performing the following temporal difference (TD) updates:
\begin{equation}
\label{q_update}
Q(s_{t},a_{t}) \leftarrow Q(s_{t},a_{t}) + \alpha[r + \gamma \max_{a_{t}^{'}} Q(s_{t}^{'},a_{t}^{'}) - Q(s_{t},a_{t})]
\end{equation}
where Q(s,a) is the state-action value given state $s$ and action $a$, $\alpha$ is the learning rate and $\gamma$ is the discounted coefficient. At each step, a random batch of experiences are sampled from the replay buffer to update the network's parameters~\cite{mnih2013playing, mnih2015human, wang2016sample, silver2016mastering}. The ER breaks the temporal correlations and increases data usage, stabilizing the training process. To further make good use of the experiences, instead of uniformly selecting experiences from the buffer, prioritized experience replay~\cite{schaul2015prioritized} samples transitions with high expected learning process more frequently. The probability of selecting a experience is determined by the relative magnitude of the TD error. However, the above methods do not fully take into consideration the intrinsic characteristics of the transition distribution in state space. The agent in RL often explores the environment from the initial state when failed and thus the transitions near the initial state appears more frequently, which makes the distribution of transitions in state space intrinsically skew as shown in Fig.~\ref{fig:unbalance}. Directly applying uniform sampling without considering the transition distribution is improper and ineffective. Actually, uniform sampling is essentially like occurrence frequencies-based sampling, which means that the $Q(s,a)$ in state space densely distributed with transitions updates more frequency while the $Q(s,a)$ in state space sparsely distributed with transitions rarely perform updates. In general, during the learning process, the more updates the $Q(s,a)$ performs, the less uncertainty the $Q(s,a)$ has. Therefore, uniform random sampling can cause that the $Q(s,a)$ in state space rarely distributed with transitions tend to be uncertainty and that some of the $Q(s,a)$ updates in state space densely distributed with transitions are likely redundant and unnecessary. Moreover, it can be seen from Eq.\eqref{q_update} that $Q(s,a)$ is updated in an iterative way and they depend on each other. The uncertainty in one $Q(s,a)$ may result in an inaccurate estimation of others.

Motivated by the observations above, we in this paper investigate how the sampling method takes transition distribution into consideration can make experience replay more efficient and effective than uniform random sampling. The key intuition is that the transition distribution in state space is intrinsically skewed and some transitions are redundant. The transitions used to update these state-action values with more uncertainty should be replayed more times to eliminate the uncertainty rather than the ones with more occurrences in uniform sampling. On the opposite, the transitions used to update these state-action values with less uncertainty should be replayed less times to reduce unnecessary updates rather that the ones with less occurrences in uniform sampling. Specifically, we propose a novel state distribution-aware sampling method to balance the replay times of transitions with different occurrence frequencies. In this way, the replay times of transitions are dependent on the uncertainty of $Q(s,a)$ instead of the occurrence frequencies, which will improve the effectiveness of TD updates and thus speed up the convergence of the learning process.

The main contributions of this work are summarized as follows:
\begin{enumerate}
  \item We propose a novel state distribution-aware sampling method for deep Q-learning, which takes into account both the occurrence frequencies of transitions and the uncertainty of the corresponding $Q(s,a)$. It balances the replay frequencies for transitions with a skew distribution and can improve the effectiveness of the experience replay.
  \item We present a static hash based method to cluster the transitions and attain the transition distribution in state space, which can deal with continuously coming transitions efficiently and effectively alleviate the problem of the high-dimensional state space.
  \item Extensive experiments are conducted on both classic control and Atari 2600 games tasks based on OpenAI gym platform and the experimental results show that the proposed sampling method can speed up the convergence and improve the policy's quality.
\end{enumerate}

\section{Our Approach}
\label{sec:1}
\subsection{Problem Formulation}
In this paper, we focus on how to accurately estimate the state-action value function by learning from the online collected transitions and simultaneously keep faster convergence and sample efficiency. Formally, consider a reinforcement learning task with the experience replay buffer $\mathcal{D} = \{(s_{t},a_{t},r_{t},s_{t}^{'})\}_{t=1}^{n}$, where $(s_{t},a_{t},r_{t},s_{t}^{'})$ is a transition tuple. Let $Q(s,a;\theta)$ denotes the deep Q-learning network needed to learn. At each learning step, a batch of transitions is iteratively sampled based on their sampling probability $p$ from the replay buffer $\mathcal{D}$, and then used to update $Q(s,a;\theta)$ via TD learning as shown in Eq.~\eqref{q_update}.

Our objective in this paper is to assign proper sampling probability $p_{i}$ to each transition $i$ to make experience replay more effective. Instead of equally assigning the probability of sampling a transition $p_{i} = \frac{1}{n}$ in uniform sampling, our main idea is that the sampling method should not only consider the occurrence frequencies of transitions but also take into account the $Q(s,a;\theta)$ uncertainty.
\begin{figure}[t]
\centering
    \begin{subfigure}[b]{0.32\textwidth}
        \includegraphics[width=\textwidth]{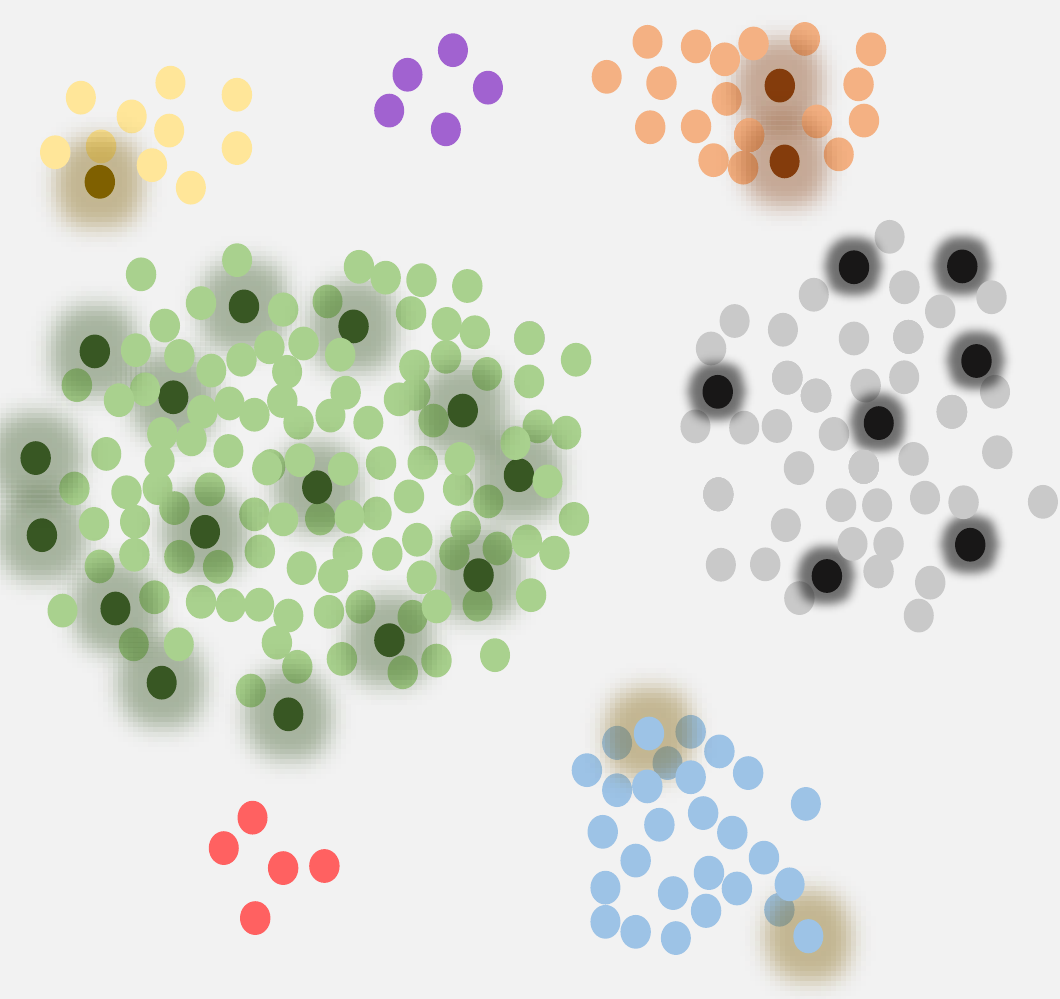}
        \caption{uniform random sampling}
    \end{subfigure}
    \begin{subfigure}[b]{0.32\textwidth}
        \includegraphics[width=\textwidth]{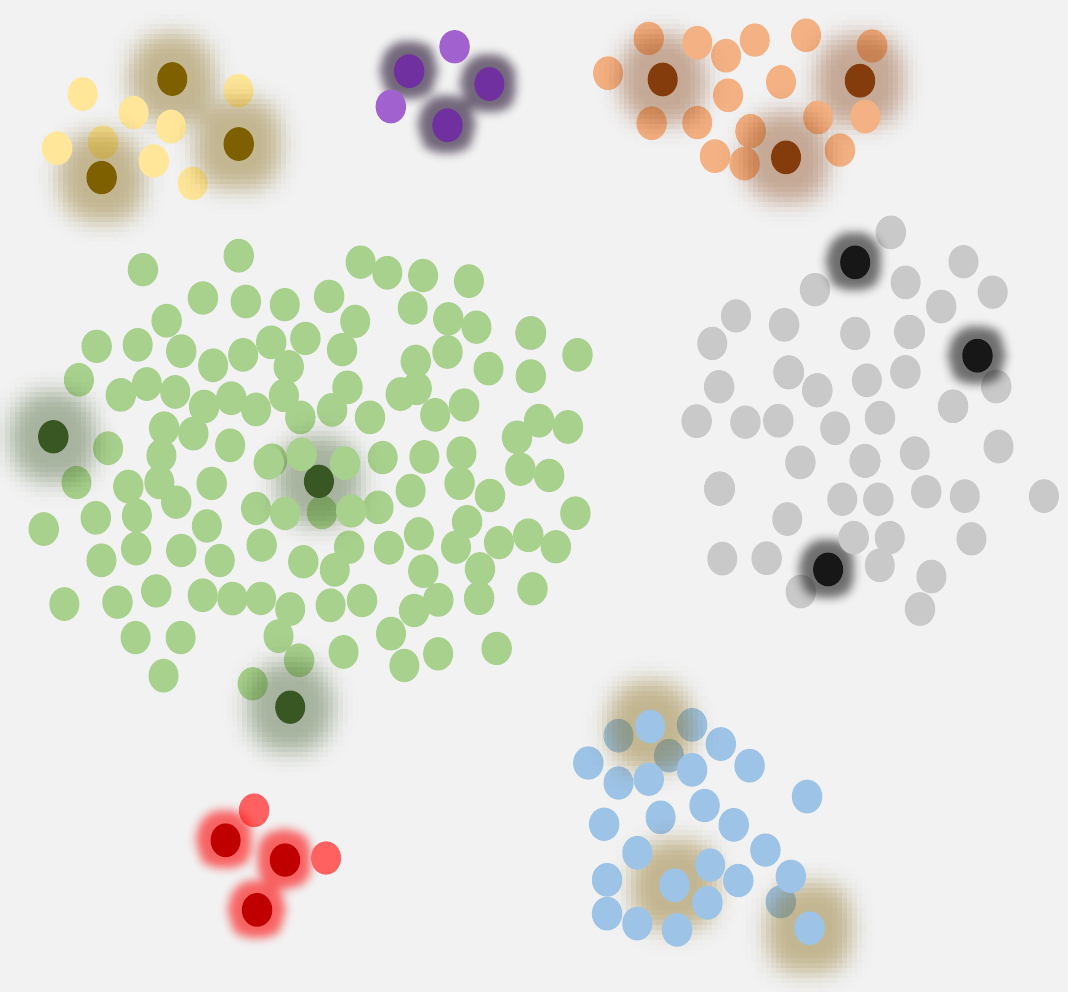}
        \caption{\scriptsize{equally cluster-based sampling}}
    \end{subfigure}
    \begin{subfigure}[b]{0.32\textwidth}
        \includegraphics[width=\textwidth]{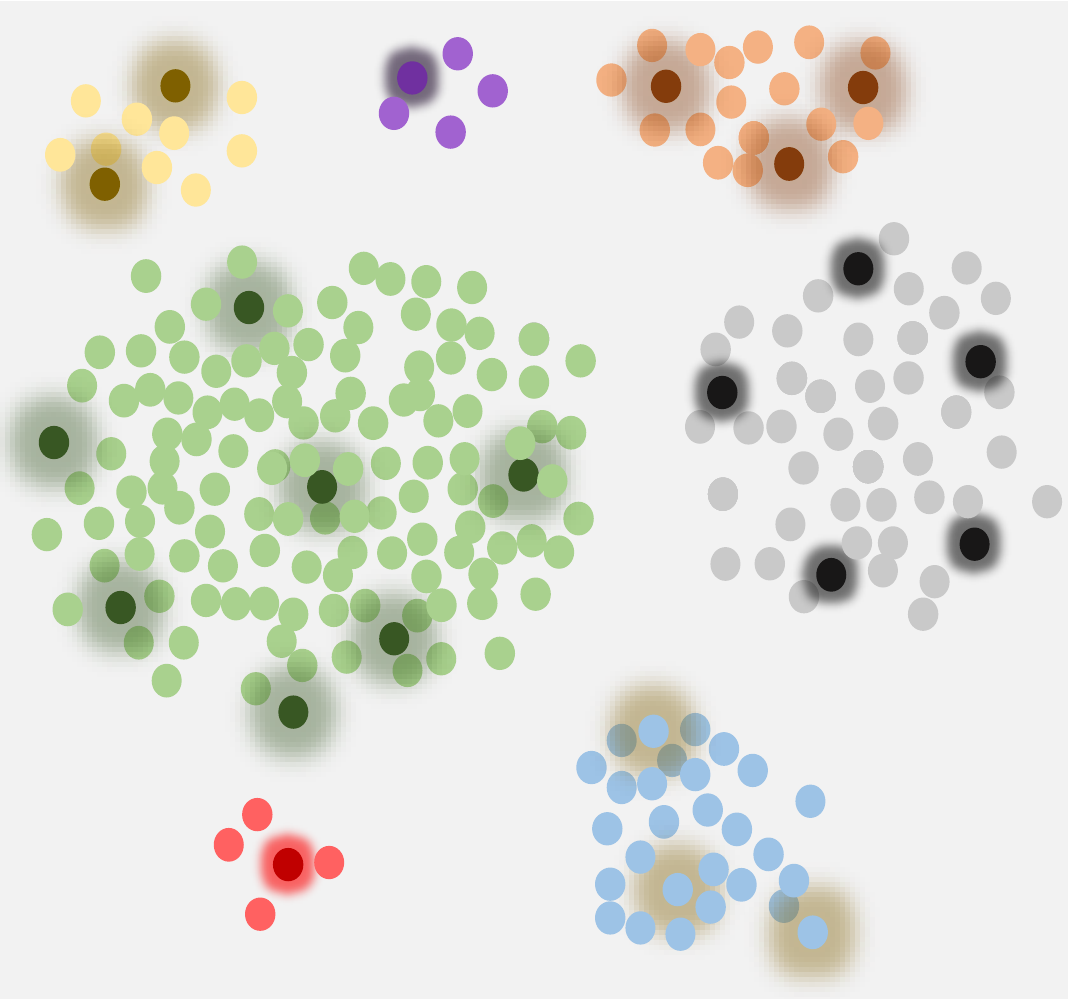}
        \caption{Our approach}
    \end{subfigure}
   \caption{Illustration of the simplified results of uniform random sampling (a), equally cluster-based sampling (b) and state distribution-aware sampling (c), the darker dots represent the transitions selected from experience replay. The state distribution-aware sampling is more reasonable.}
\label{fig:sampling_diff}
\end{figure}
\subsection{State Distribution-aware Sampling for Experience Replay}
\label{sec:2}
In this section, we present our state distribution-aware sampling method to solve the imbalanced transition distribution problem. Considering the fact that a transition $(s_{t},a_{t},r_{t},s_{t}^{'})$ is only connected to the $Q(s_{t},a_{t})$ update, we do not need get the transition distribution over the 4-tuples and instead we need attain the distribution of the first state property of 4-tuples in state space. In order to approximately attain the distribution, we discretize the state space and employ a hash based clustering method (details presented in section~\ref{sec:3}) to cluster all the transitions in the replay buffer $\mathcal{D}$ into $k$ different clusters according to their first state properties. To alleviate the $Q(s,a;\theta)$ uncertainty, the replay times of transitions in one cluster with less transitions should be increased and the replay times of transition in one cluster with more transitions should be decreased. The simplest way to solve this issue is to select equal transitions from each cluster to replay, i.e. equally cluster-based sampling, in which the probability of sampling transition $i$ can be defined as:
\begin{equation}
p_{i}=\frac{1}{k*num_{i}}
\end{equation}
here $num_{i}$ denotes the number of transitions in the cluster the transition $i$ belongs to. In this way, the uncertainty problem can be mitigated but it could cause some issues where the transitions in one cluster with few transitions will be seriously over-sampled while the transitions in one cluster with huge transitions will be seriously under-sampled to discard some potentially useful ones.

Therefore, we take into consideration both the occurrence frequencies of transitions and the $Q(s,a;\theta)$ uncertainty, and introduce a distribution-aware sampling method that interpolates between uniform random sampling and equally cluster-based sampling. Concretely, we define the probability of sampling transition $i$ as
\begin{equation}
\label{tradeoff_sel}
p_{i}=\beta*\frac{1}{n} + (1-\beta)*\frac{1}{k*num_{i}}
\end{equation}
here $\beta\in[0,1]$ is the hyper parameter which balances between uniform random sampling and equally cluster-based sampling. With $\beta=0$, the Eq.~\eqref{tradeoff_sel} corresponds the uniform case. Otherwise, the Eq.~\eqref{tradeoff_sel} becomes the equally cluster-based sampling with $\beta=1$. Seen from Eq.~\eqref{tradeoff_sel}, the distribution-aware sampling method can relatively reduce the $Q(s,a;\theta)$ updates in state space densely distributed with transitions and relatively increase the $Q(s,a;\theta)$ updates in state space sparsely distributed with transitions, thus achieving the effect that balances the replay frequencies of transitions with skew distribution.

Next, we present a toy example to help understand the difference of the above sampling methods. The visualized results are conducted on the BattleZone task in Atari 2600 game via  PCA and t-SNE\cite{maaten2008visualizing}.
As is shown in Fig.~\ref{fig:sampling_diff}, the transitions in different clusters are represented in a different color. The darker and bigger points are the transitions selected to update the action-value function $Q(s,a;\theta)$.
It can be seen from Fig.~\ref{fig:sampling_diff} (a) that  most of the transitions sampled belong to the clusters with huge transitions while transitions in some clusters with few transitions are not selected. From Fig.~\ref{fig:sampling_diff} (b), it can be seen that equally cluster-based sampling will select equal transitions from each cluster. Different from the two methods above, our approach shown in Fig.~\ref{fig:sampling_diff} (c) tends to select transitions from every cluster and simultaneously considers the number of transitions in every cluster, which is more reasonable.

\subsection{Hash based Online Clustering for Transitions}
\label{sec:3}
In RL, the number of transitions in experience replay buffer is usually huge and the new transitions are continuously generated. Further, the representation of each state is high-dimensional, especially for Atari games. Therefore, it is impractical to adopt the standard k-means method to implement the online clustering~\cite{mahajan2009planar,hartigan1979algorithm}. In order to deal with continuously coming transitions efficiently and improve the speed of clustering, we use locality-sensitive hash (LSH) to convert continuous, high-dimensional data to discrete hash codes rather than directly and repeatedly clustering them. LSH provides a popular method to query nearest neighbors based on certain similarity metrics~\cite{bloom1970space}. Specifically, we employ the SimHash~\cite{andoni2006near}, an effective implementation of LSH, to retrieve a lower dimension binary code of the state $s_{t}$. The distant states will be projected to different hash codes and while the similar states will be mapped to the identical hash codes. The unique hash code can be regarded as the identity of one cluster with transitions. In this way, the transition distribution in state space can be approximately attained according to the hash table counts.

In addition, state representation is also important for clustering. Here we use deep convolutional features to give a better description of each state than handcrafted features for Atari games. In practise, we directly leverage the output of the 3rd convolutional layer as state features in deep Q-learning network.

\subsection{Deep Q-learning with State Distribution-aware Sampling}
We integrate our approach into a RL agent, based on the state-of-art Double DQN algorithm~\cite{van2016deep}. Our principal modification is to replace the uniform random sampling with state distribution-aware sampling. The details of this algorithm can be seen from Algorithm~\ref{alg:outline}. When a transition is generated, we will firstly compute the hash codes of the state. The transitions with identical hash codes belong to the same cluster. Further, we update the hash table counts. In this way, we can online cluster transitions into different clusters and obtain the number of transitions in each cluster. Then, the probability of sampling each transition will be updated via Eq.~\eqref{tradeoff_sel}. Finally a batch of transitions will be sampled based on their current probabilities to perform $Q(s,a;\theta)$ updates via Eq.~\eqref{q_update}.
\begin{algorithm}
	\caption{deep Q-learning with state distribution-aware sampling}
	\label{alg:outline}
	\begin{algorithmic}[1]
    \STATE Initialize replay buffer $\mathcal{D}$, action-value function $Q$ with random weights $\theta$, target action-value function $\hat{Q}$ with weights $\theta^{-}$, hash table with a size of $k$ and counts $h(\cdot)\equiv0$.
    \FOR{$episode = 1 \to M$}
    \FOR{$t = 1 \to T$}
    \STATE Given current state $s_{t}$, select action $a_{t}$ with $\varepsilon-greedy$ policy~\cite{mnih2013playing}.
    \STATE Execute $a_{t}$ and observe reward $r_{t}$ and next state $s_{t+1}$.
    \STATE Store the transition tuple $(s_{t}, a_{t}, r_{t}, s_{t+1})$ into $\mathcal{D}$.
    \STATE Compute the hash codes $c$ of the state $s_{t}$ of the new transition as presented in section~\ref{sec:3}.
    \STATE Update the hash table counts as $h(c) = h(c) + 1$.
    \STATE Given $h(\cdot)$, update the probability of sampling each transition via Eq.~\eqref{tradeoff_sel}.
    \STATE Sample a batch of transitions $\{(s_{j},a_{j},r_{j},s_{j}^{'})\}_{j=1}^{B}$ from $\mathcal{D}$ based on their probabilities.
    \STATE Set $y_{j} =
        \begin{cases}
        r_{j} & \text{if episode terminates at step $j+1$;} \\
        r_{j}+\gamma \max_{a_{j}^{'}} \hat{Q}(s_{j}^{'},a_{j}^{'};\theta^{-})  &\text{otherwise.}
        \end{cases}$
    \STATE Perform a gradient descent step on $\sum_{j=1}^{B}(y_{j} - Q(s_{j}, a_{j}; \theta))^{2} $ with respect to $\theta$.
    \STATE Every $K$ steps reset $\hat{Q}= Q$.
    \ENDFOR
	\ENDFOR
	\end{algorithmic}
\end{algorithm}
\section{Experiments}
To demonstrate the effectiveness of our approach, we conduct extensive experiments on both classic control tasks and Atari 2600 games tasks based on OpenAI gym platform\cite{1606.01540}. We implement our method based on DQN\cite{hasselt2010double} in OpenAI Baselines\cite{baselines}.

\subsection{Classic Control Tasks}
\begin{figure*}[t]
\centering
    \begin{subfigure}[b]{0.32\textwidth}
        \includegraphics[width=\textwidth]{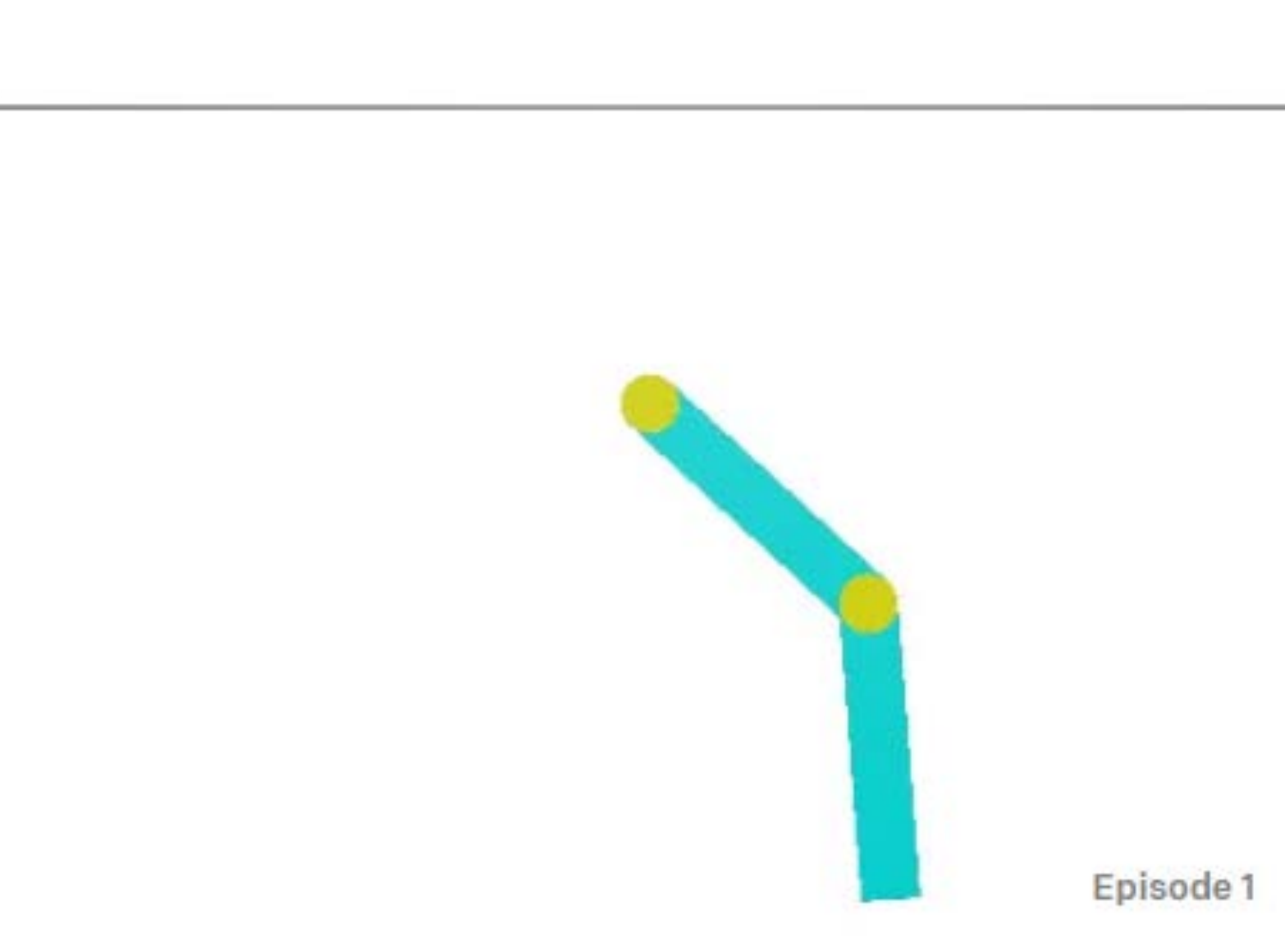}
    \end{subfigure}
    \begin{subfigure}[b]{0.32\textwidth}
        \includegraphics[width=\textwidth]{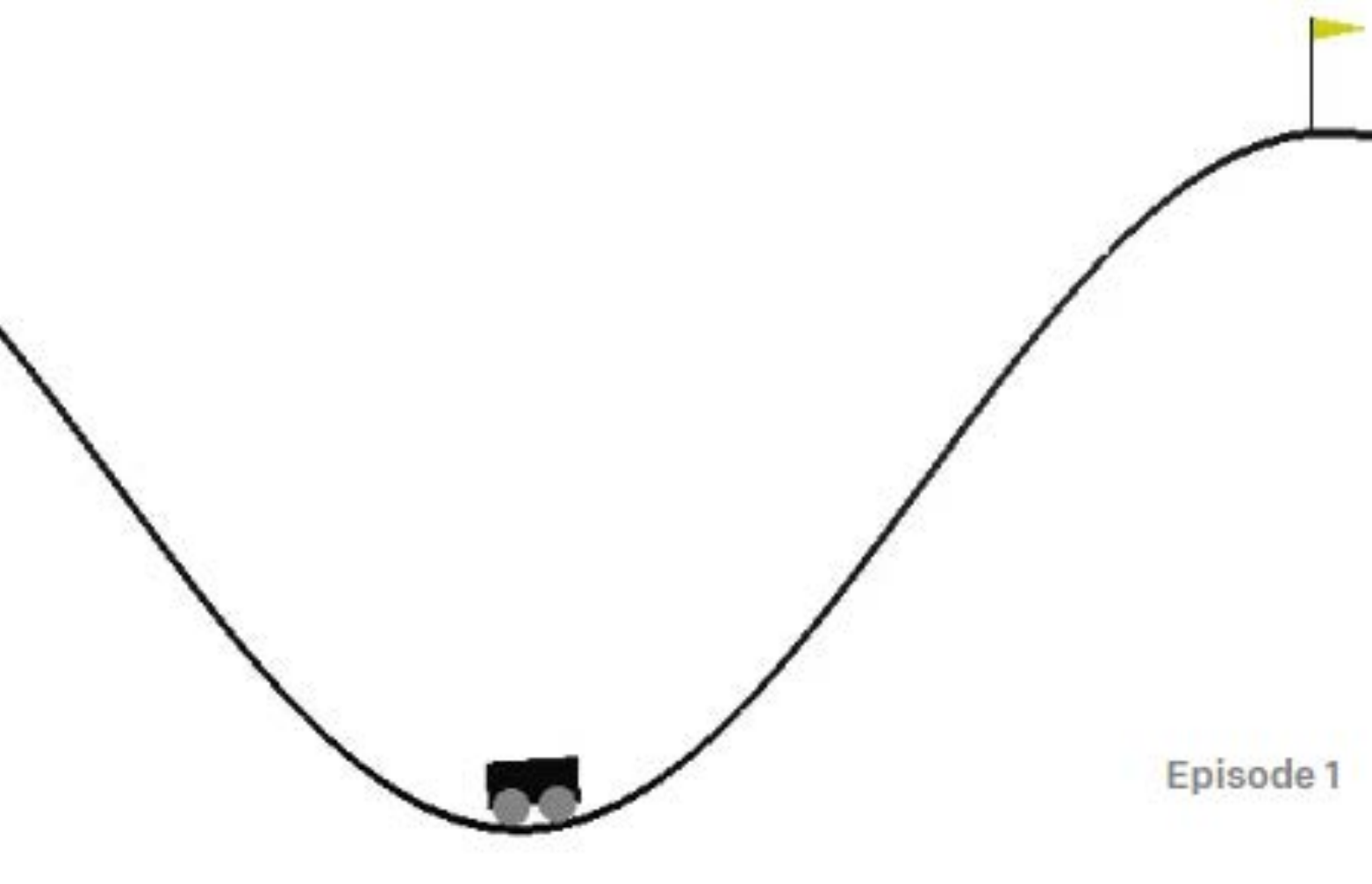}
    \end{subfigure}
    \begin{subfigure}[b]{0.32\textwidth}
        \includegraphics[width=\textwidth]{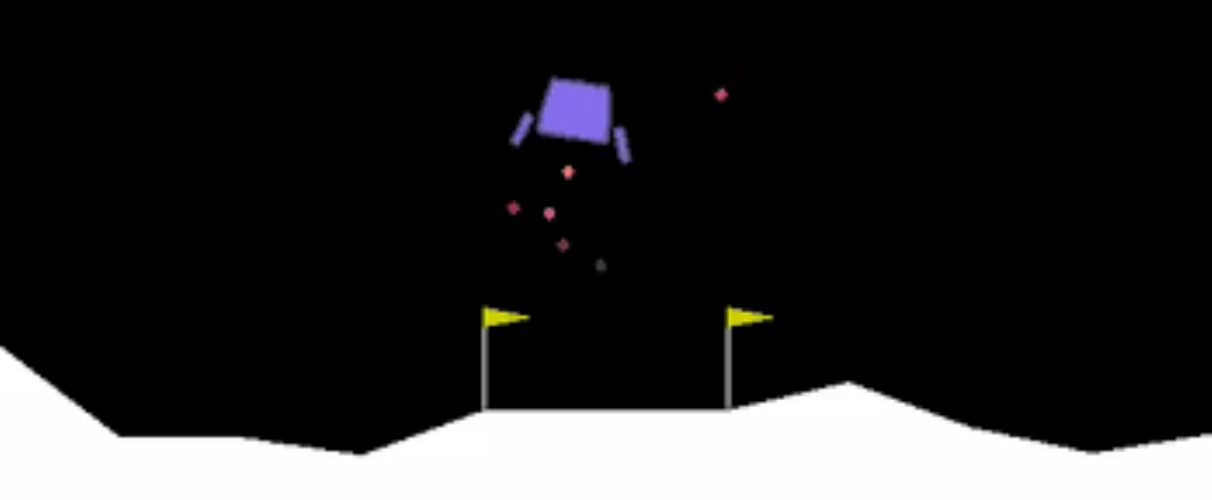}
    \end{subfigure}
    \begin{subfigure}[b]{0.32\textwidth}
        \includegraphics[width=\textwidth]{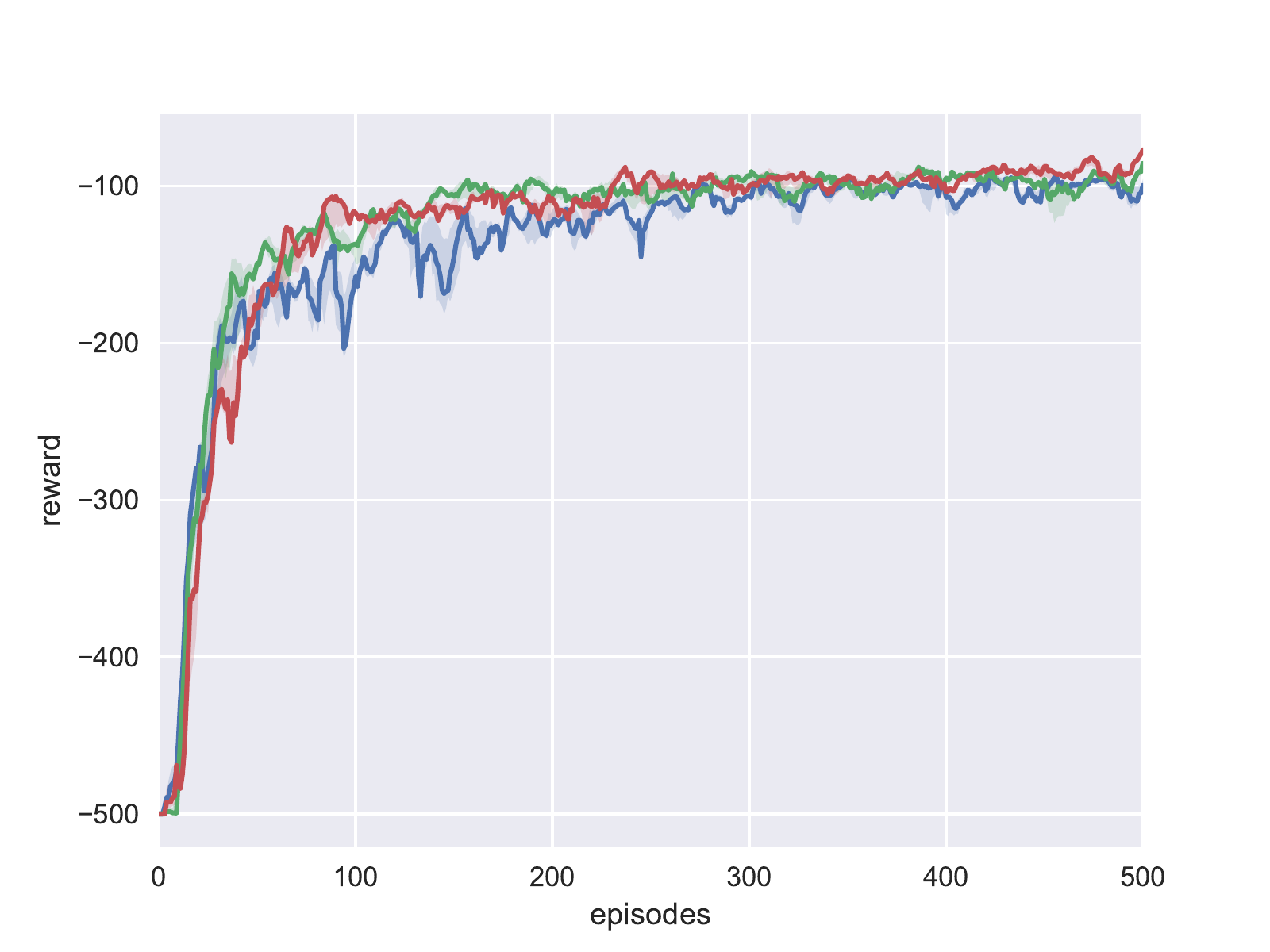}
        \caption{Acrobot}
    \end{subfigure}
    \begin{subfigure}[b]{0.32\textwidth}
        \includegraphics[width=\textwidth]{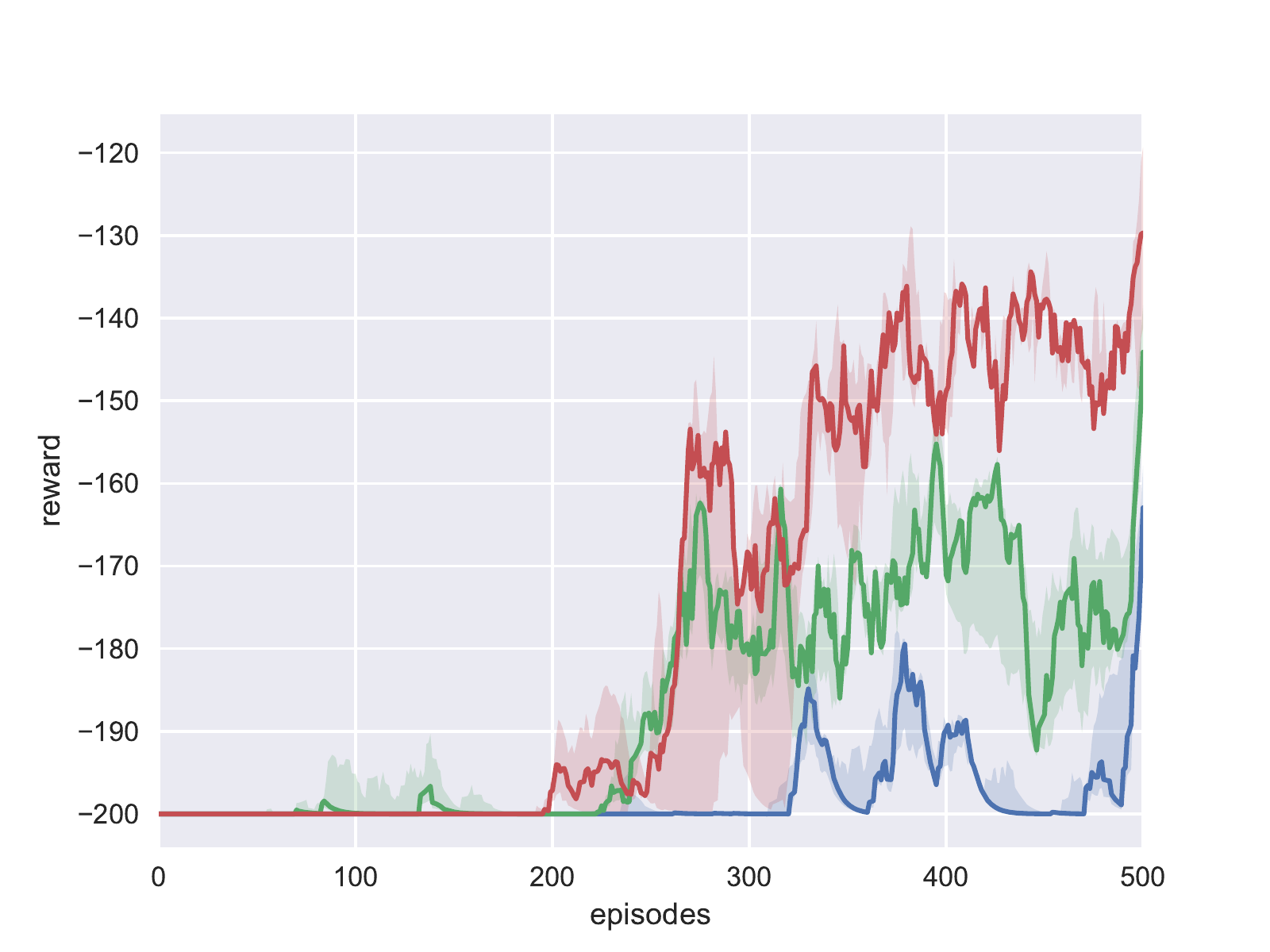}
        \caption{MountainCar}
    \end{subfigure}
    \begin{subfigure}[b]{0.32\textwidth}
        \includegraphics[width=\textwidth]{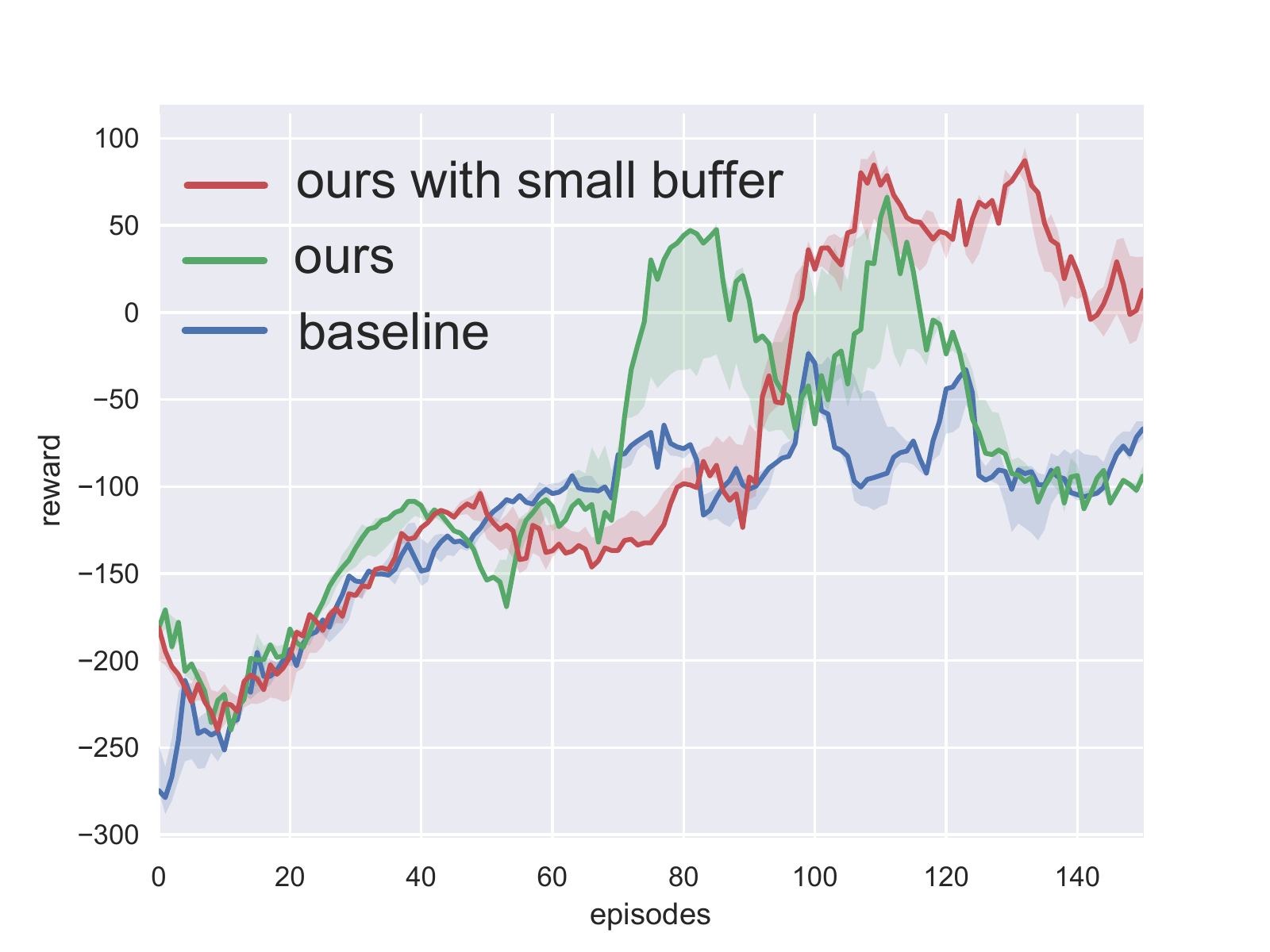}
        \caption{LunarLander}
    \end{subfigure}
   \caption{The mean average rewards curve of three classic control tasks on OpenAI Gym.
   The solid line represents the mean average rewards while the shaded area represents the range of variation.
   Our proposed approach can speed up the convergence of the learning procedure and achieve higher total rewards at the same time for all the tasks, especially for MountainCar task.}
\label{fig:classic_control_results}
\end{figure*}

The first experiment is conducted on control theory problems, and we choose Acrobot, MountainCar  and LuarLander tasks. The visualization of these tasks are shown in the first row of Fig.~\ref{fig:classic_control_results}. The state space dimension of the Acrobot task, MountainCar task and LuarLander task is only 6, 2 and 8 respectively. Therefore, we directly leverage the observation data as features for states and use standard k-means method to cluster transitions into 64 different clusters in the state space. For the sake of fairness, the other experimental settings are set to the same as~\cite{baselines}, in which the size of the reply buffer is set to $5\times10^{4}$. Additionally, we also conduct another experiment with a much smaller replay buffer with a size of $1\times10^{4}$.

The second row of Fig.~\ref{fig:classic_control_results} shows the mean average rewards curve of different experimental settings on the three classic control tasks. From Fig.~\ref{fig:classic_control_results}, it can be seen that our proposed approach, even with only $20\%$ size of the replay buffer, can speed up the convergence of the learning procedure and achieve higher total rewards after convergence at the same time for all three tasks, especially for MountainCar task.

\begin{figure*}[t]
\centering
    \begin{subfigure}[b]{0.32\textwidth}
        \includegraphics[width=\textwidth]{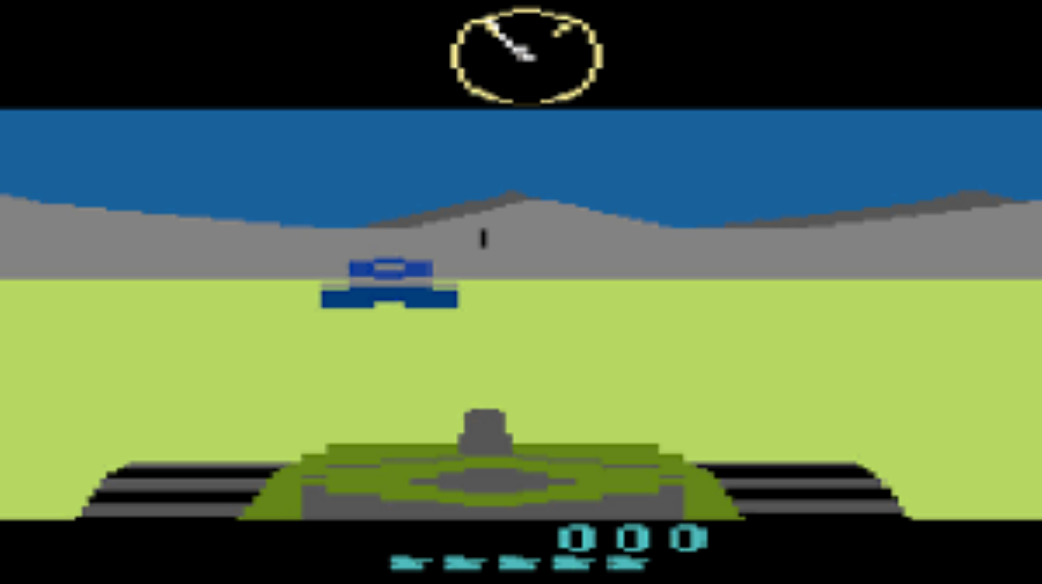}
    \end{subfigure}
    \begin{subfigure}[b]{0.32\textwidth}
        \includegraphics[width=\textwidth]{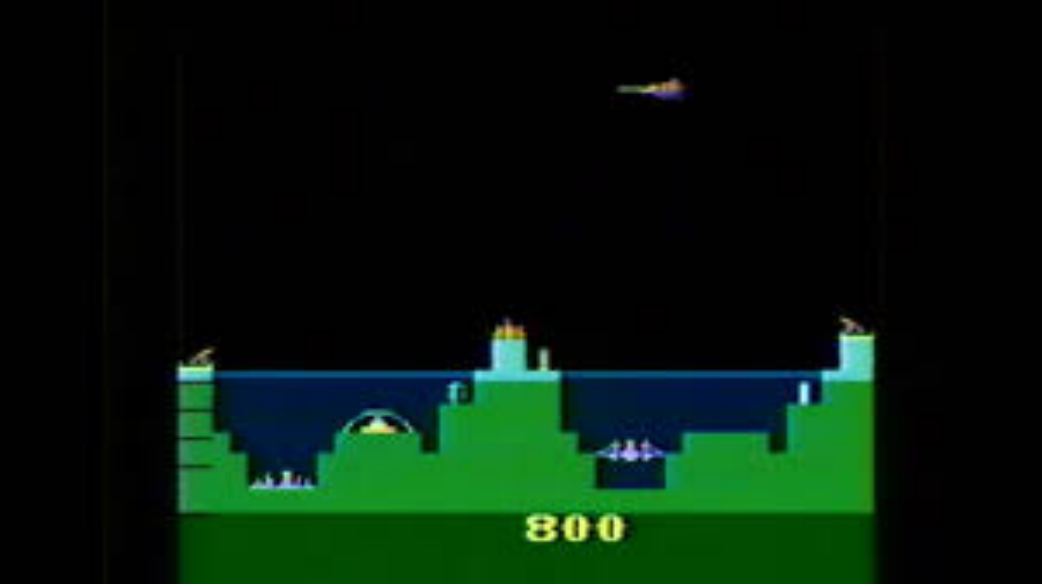}
    \end{subfigure}
    \begin{subfigure}[b]{0.32\textwidth}
        \includegraphics[width=\textwidth]{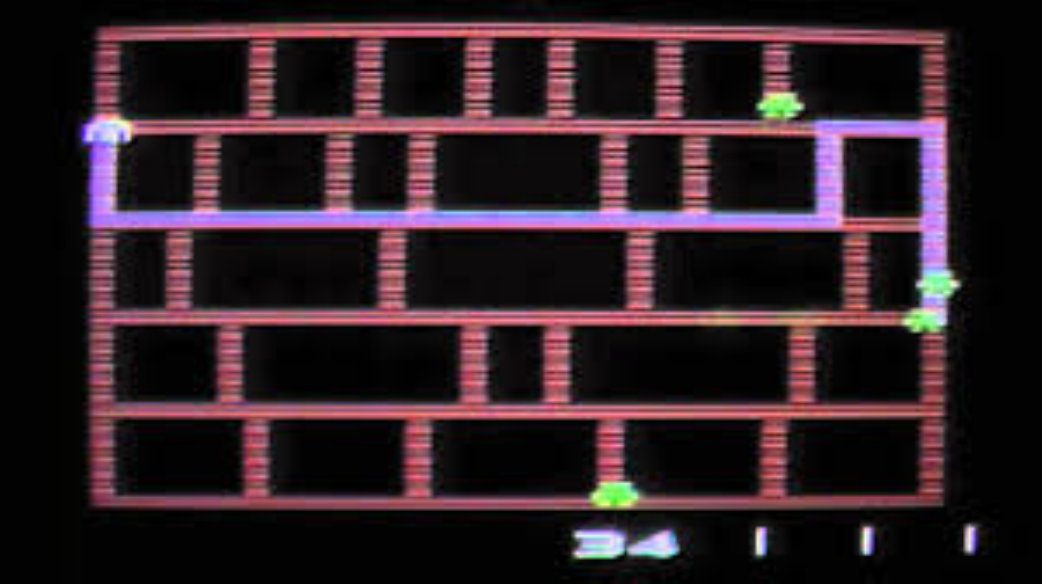}
    \end{subfigure}
    \begin{subfigure}[b]{0.32\textwidth}
        \includegraphics[width=\textwidth]{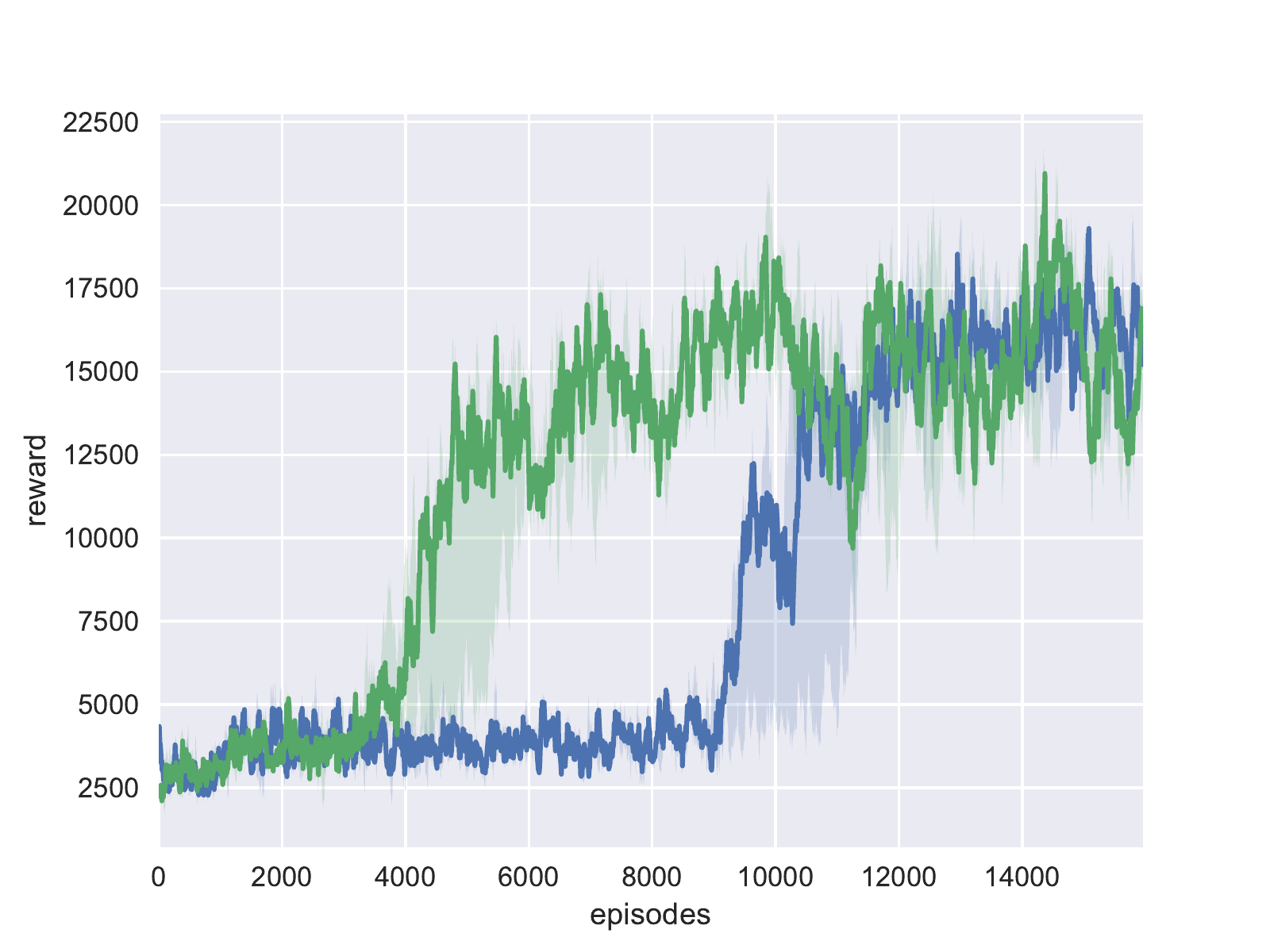}
        \caption{BattleZone}
    \end{subfigure}
    \begin{subfigure}[b]{0.32\textwidth}
        \includegraphics[width=\textwidth]{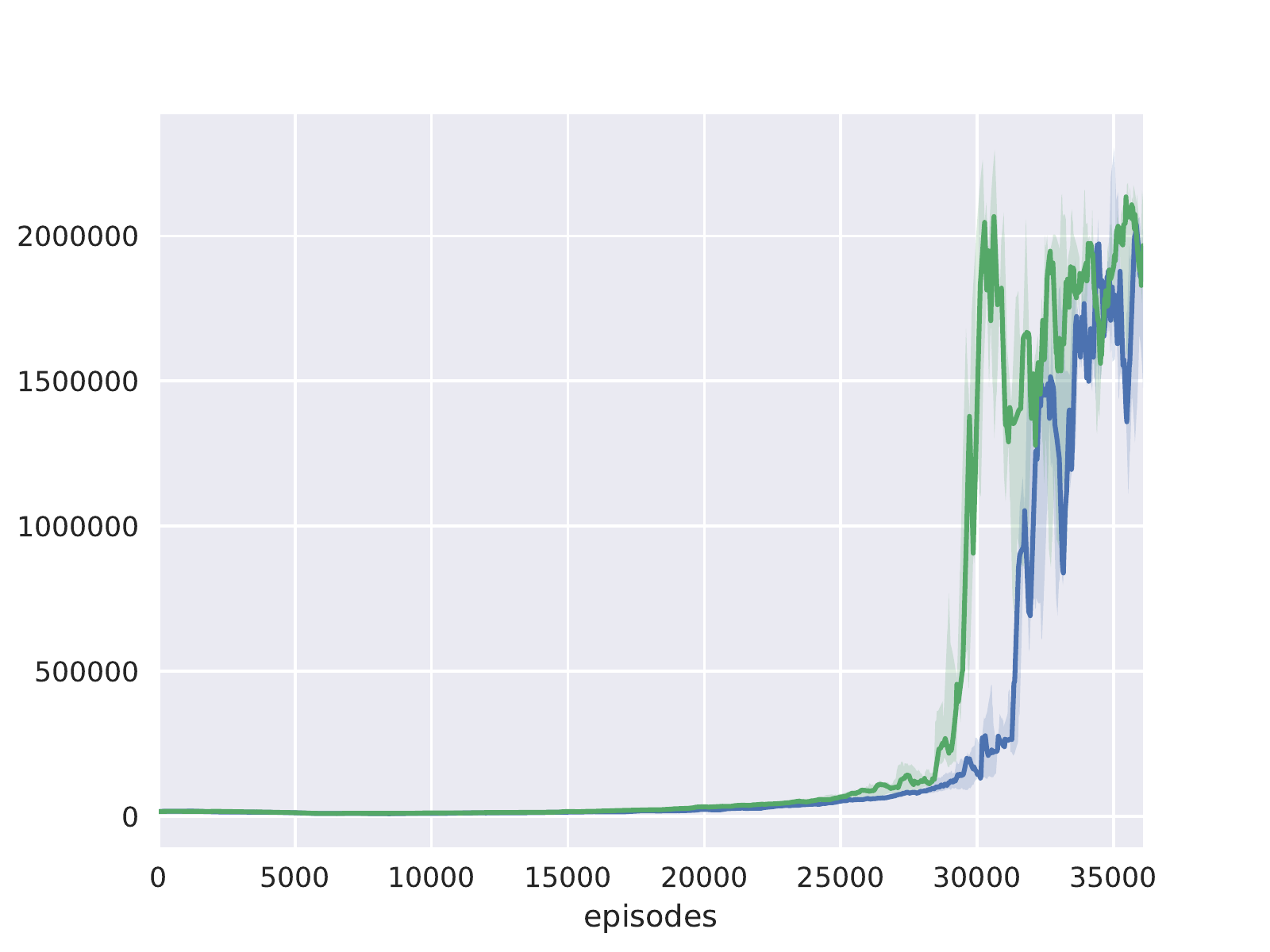}
        \caption{Atlantis}
    \end{subfigure}
    \begin{subfigure}[b]{0.32\textwidth}
        \includegraphics[width=\textwidth]{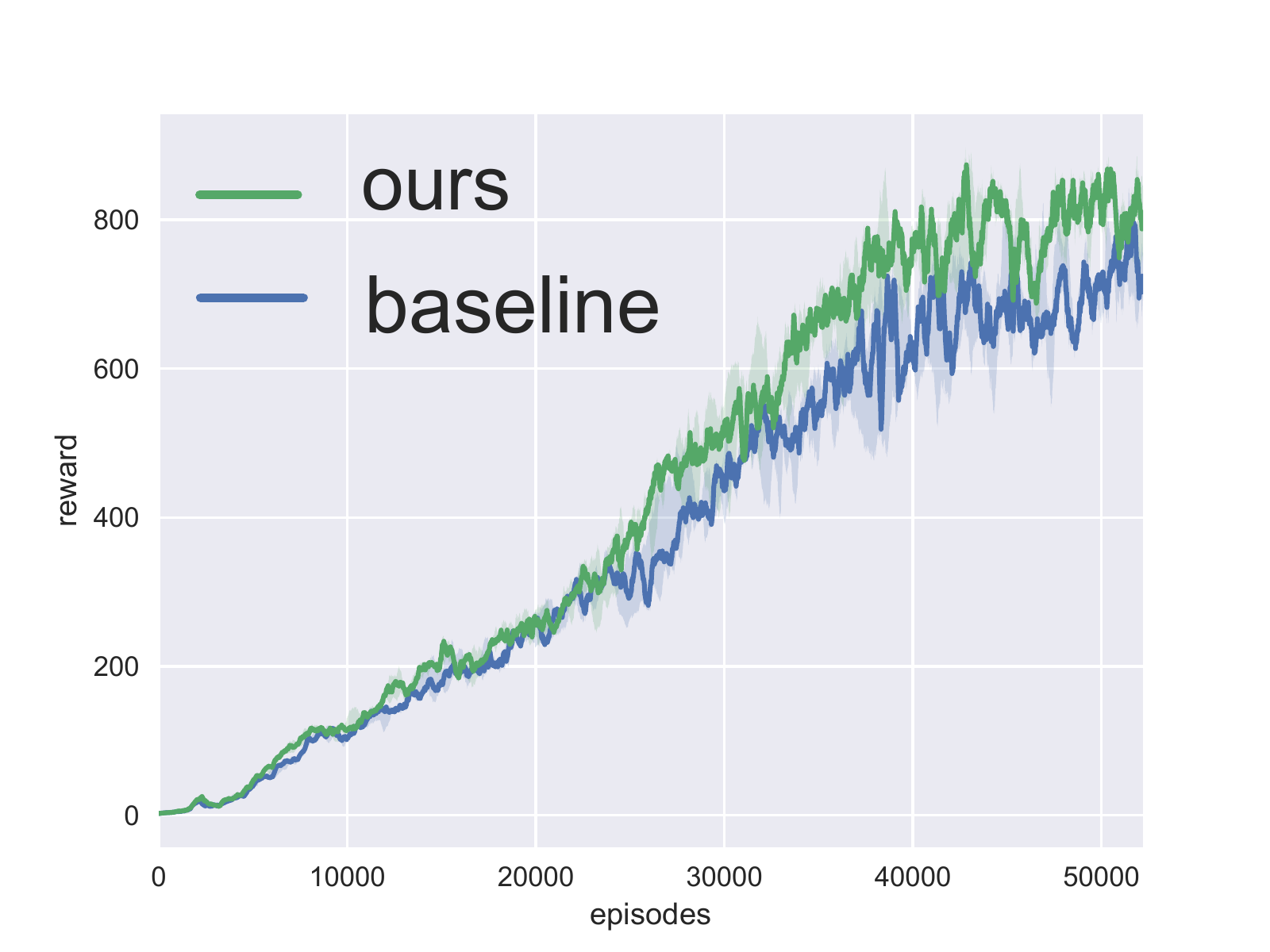}
        \caption{Amidar}
    \end{subfigure}
   \caption{The mean average rewards curve of three Atari 2600 games on OpenAI Gym between our method and the baseline DQN. The solid line represents the mean average rewards while the shaded area represents the range of variation. Our approach can achieve faster convergence and attain higher total rewards at the same time for all three tasks compared to the baseline.}
\label{fig:atari_results}
\end{figure*}

\subsection{Atari 2600 Games}
\begin{table}[h]
    \centering
        \begin{tabular}{|c|c|c|c|c|}
            \hline
                  & BattleZone  & Atlantis    & Amidar \\
            \hline
            baseline & 17834 & 1135  & 772  \\
            Ours  & \textbf{19473} & \textbf{1345}  & \textbf{838} \\
            \hline
        \end{tabular}%
    \caption{The average total rewards after convergence between our method and the baseline DQN on three Atari games. The best results are displayed in bold. Our proposed approach attains significantly higher average total rewards for all the three games than the baseline DQN, especially for the BattleZone game.
}
    \label{tab:atari_rewards_table}%
\vspace{-0.8\baselineskip}
\end{table}%
We conduct the second experiment on Arcade Learning Environment (ALE)~\cite{bellemare2013arcade} and choose BattleZone, Atlantis and Amidar from Atari 2600 video games as experimental targets. The screenshots of these games are displayed in the first row of Fig.~\ref{fig:atari_results}. It can be seen that these task are more difficult that the classic control problems above. The observations of these tasks are high-dimensional raw images. For instance, the default DQN~\cite{baselines} for  Atari games uses four consecutive frames to represent the observation, the dimension of which is 28224 (84x84x4). The raw images contain some redundant information. To efficiently attain the transition distribution in state space, we employ the output of the 3rd convolutional layer in deep-Q learning network as state's features to perform transitions clustering. Here, all the transitions in reply buffer are clustered into 128 clusters via the hash based clustering method presented in section~\ref{sec:3}. For the sake of fairness, the other experimental settings are set to the same as~\cite{baselines}, in which the size of the reply buffer is set to $10^{6}$.

The average total rewards after convergence for three tasks are shown in Table~\ref{tab:atari_rewards_table}. It can be seen from Table~\ref{tab:atari_rewards_table} that our proposed approach attains significantly higher average total rewards for all the three games than the baseline DQN. Particularly, Our method attains 1345 total rewards in the Atlantis game, with about 18.5\% higher than the baseline. Moreover, our method achieves about 9.2\% and 8.5\% improvement on BattleZone and Amidar task respectively.

The second row of Fig.~\ref{fig:atari_results} shows the mean average rewards curve of three Atari games between our method and the baseline. It can be seen from Fig.~\ref{fig:atari_results} that our approach achieves faster convergence and attains higher total rewards at the same time for all three tasks compared to the baseline. Particularly, in the BattleZone game, Our approach begins to converge at about 5000 episodes while the baseline DQN begins to converge at about 10000 episodes. In conclusion, our method can significantly reduce the time taken to converge and improve the policy's quality compared to baseline.

\section{Conclusion}
In this paper, we have proposed a novel state distribution-aware sampling method for deep Q-learning to solve the imbalanced transition distribution problem. The sampling method takes into consideration both the occurrence frequencies of transitions and the $Q(s,a)$ uncertainty, balancing the replay times of transitions with different occurrence frequencies. Extensive experiments on both classic control tasks and Atari 2600 games have shown that our approach can make experience replay more effective and efficient compared to the baseline DQN.

%


\bibliographystyle{spbasic}      
\bibliography{sdaer}   
%
\end{document}